%% file: main.tex
\documentclass[conference]{IEEEtran}
\IEEEoverridecommandlockouts

\usepackage[ruled,linesnumbered]{algorithm2e}
\usepackage{multirow}

\usepackage{cite}
\usepackage{amsmath,amssymb,amsfonts}
\usepackage{algorithmic}
\usepackage{graphicx}
\usepackage{textcomp}
\usepackage{xcolor}

\usepackage{caption}
\usepackage{subcaption}
\def\BibTeX{{\rm B\kern-.05em{\sc i\kern-.025em b}\kern-.08em
    T\kern-.1667em\lower.7ex\hbox{E}\kern-.125emX}}
    
\newcommand{\rev}[1]{#1}
\usepackage{tikz}
\usepackage{textcomp}
\usepackage[doipre={DOI:~}]{uri}
\usepackage{lipsum}
\newcommand\copyrighttext{%
  \footnotesize \textcopyright 2022 IEEE. Personal use of this material is permitted.  Permission from IEEE must be obtained for all other uses, in any current or future media, including reprinting/republishing this material for advertising or promotional purposes, creating new collective works, for resale or redistribution to servers or lists, or reuse of any copyrighted component of this work in other works. 

  Published as a conference paper at the IEEE 2021 VLSI-SoC Conference.  ~\doi{10.1109/VLSI-SoC53125.2021.9606986} }
\newcommand{\copyrightnotice}{%
\begin{tikzpicture}[remember picture,overlay]
\node[anchor=south,yshift=10pt] at (current page.south) {\fbox{\parbox{\dimexpr\textwidth-\fboxsep-\fboxrule\relax}{\copyrighttext}}};
\end{tikzpicture}%
}
\begin{document}
\bstctlcite{IEEEexample:BSTcontrol}

\title{\vspace{-0.3cm}Adaptive Random Forests for Energy-Efficient Inference on Microcontrollers}

\author{\IEEEauthorblockN{Francesco Daghero$^*$, Alessio Burrello$^\dagger$, Chen Xie$^*$, Luca Benini$^\dagger$, Andrea Calimera$^*$, Enrico Macii$^*$,\\Massimo Poncino$^*$, Daniele Jahier Pagliari$^*$}
\IEEEauthorblockA{$^*$Politecnico di Torino, Turin, Italy, name.surname@polito.it\\
$^\dagger$University of Bologna, Bologna, Italy, name.surname@unibo.it}
}

\IEEEoverridecommandlockouts
\IEEEpubid{\makebox[\columnwidth]{978-1-6654-2614-5/21/\$31.00~\copyright2021 IEEE \hfill}\hspace{\columnsep}
\makebox[\columnwidth]{ }}

\maketitle
\copyrightnotice%

\begin{abstract}
\input{sections/00_abstract}
\end{abstract}


\section{Introduction}\label{sec:intro}
\input{sections/01_introduction}

\section{Background}\label{sec:background}\label{sec:rf}
\input{sections/02_background}

\section{Adaptive Random Forests}
\input{sections/03_method}

\section{Experimental Results}\label{sec:results}
\input{sections/04_results}

\section{Conclusions}
\input{sections/05_conclusions}

\tiny

\end{document}

%% file: sections/00_abstract.tex
Random Forests (RFs) are widely used Machine Learning models in low-power embedded devices, due to their hardware friendly operation and high accuracy on practically relevant tasks.
The accuracy of a RF often increases with the number of internal weak learners (decision trees), but at the cost of a proportional increase in inference latency and energy consumption.
Such costs can be mitigated considering that, in most applications, inputs are not all equally difficult to classify. Therefore, a large RF is often necessary only for (few) hard inputs, and wasteful for easier ones.
In this work, we propose an \textit{early-stopping} mechanism for RFs, which terminates the inference as soon as a high-enough classification confidence is reached, reducing the number of weak learners executed for easy inputs.
The early-stopping confidence threshold can be controlled at runtime, in order to favor either energy saving or accuracy.
We apply our method to three different embedded classification tasks, on a single-core RISC-V microcontroller, achieving an energy reduction from 38\% to more than 90\% with a drop of less than 0.5\% in accuracy.
We also show that our approach outperforms previous adaptive ML methods for RFs.

%% file: sections/01_introduction.tex
Machine Learning (ML) inference is at the core of many emerging Internet of Things (IoT) applications, ranging from time-series processing to computer vision~\cite{Samie2019}.
In recent years, a lot of research has been devoted to optimize ML models to enable the execution of inference tasks directly on IoT end-nodes, with the goal of improving the standard cloud-centric approach on several non-functional metrics~\cite{Zhou2019}. Specifically, besides a lower and more predictable latency in presence of unstable connectivity, and an enhanced data privacy, performing inference on end-nodes often results in a higher energy efficiency too, by avoiding the transmissions of large amounts of raw data through a power-hungry wireless link~\cite{Zhou2019}.

However, this potential can only be realized if the complexity of ML models is made compatible with the limited compute and memory resources and extremely tight energy budgets of IoT nodes, most of which are based on Microcontrollers (MCUs). This is particularly challenging for Deep Learning (DL) approaches, which despite their state-of-the-art accuracy on many tasks are often too heavy for MCUs, even after applying multiple optimizations~\cite{Jacob2018}. Fortunately, non-deep models are often sufficient for simple tasks, yielding comparable results with far lower complexity.
In particular, decision-tree-based models have found success both in academia and industry for ultra-low-power applications such as human activity recognition (HAR) and seizure detection~\cite{fan2013,stsensor,Donos2015}, thanks to the fact that inference is based on a relatively small number of compare and branch operations, and that their memory footprint is compact.
In particular, Random Forests (RFs)~\cite{breiman2001}, i.e.,  ensembles of decision trees grown on random samples of the training data,
typically reach a significantly higher accuracy than individual trees, 
\rev{yet with a much lower complexity compared to DL solutions. For instance, the deep model proposed in~\cite{ecgAnomaly} for Electrocardiogram anomaly detection requires around 200k arithmetic operations and the storage of as many parameters. Instead, the baseline RF used for our experiments in Section~\ref{sec:results} achieves a comparable accuracy with around 2k parameters and less than 1k operations.
Still, on single core MCUs, RFs inference time and energy consumption are linearly proportional to}
the number of trees, even though executing the entire RF may only be useful for particularly difficult inputs. Intuitively, in fact, most trees will agree on their predictions for an easy input. In those cases, executing a subset of the ensemble could suffice, as it is unlikely that the skipped trees would overturn the global (averaged) RF output.

In this work we concretize this intuition, proposing an \textit{early-stopping} mechanism for RFs, which terminates the execution as soon as a high-enough prediction confidence is reached, with the goal of reducing the average energy consumption for inference. While similar \textit{adaptive inference} approaches have been recently tested for deep learning~\cite{Daghero2020,Park2015,JahierPagliari2018a,Tann2016,Panda2016,JahierPagliari2020b}, their use with RFs is much less explored.
The few existing solutions are either very complex~\cite{Gao2011} or limited to 2-class problems~\cite{Wang2018}. Moreover, they have never been deployed on embedded devices for energy efficiency. To our knowledge, ours is the first approach to consider these practical aspects in a MCU deployment scenario, thanks to an early-stopping solution featuring i) a runtime-controllable knob to tune the energy vs accuracy trade-off and ii) minimal latency and energy overheads compared to the execution of a single tree.
We test our approach on three different tasks, i.e., HAR, heart failure detection and gesture recognition, on a popular single-core RISC-V MCU, obtaining an energy reduction ranging from 38\% to 90\% with less than 0.5\% accuracy drop.

%% file: sections/02_background.tex
Decision Trees (DTs) implement classification and regression tasks by assigning different output predictions to different axis-parallel rectangular partitions of the feature space. This is obtained comparing, in each node, one of the input's features with a threshold, and then recursively executing either the left or right sub-trees based on the result.
The inference starts at the tree's \textit{root}, and ends at one of the \textit{leaves}, which contain the output prediction corresponding to the input. 

Describing the training (or growth) procedure of DTs is out of our scope. Interested readers can refer to~\cite{dt_book}.  Here, we focus on computational aspects of the inference. Algorithm~\ref{alg:dt} shows a high-level inference pseudo-code, where
$T$ denotes the tree.
Evidently, the time complexity of the main while loop is $O(D)$ where $D$ is the \textit{depth} (i.e., the number of levels) of the tree, while the memory complexity grows with the total number of nodes, i.e., $O(2^{D})$, for an \textit{unpruned} tree~\cite{dt_book}.

\begin{algorithm}[ht]
\SetAlgoLined
\small
$n = \mathrm{Root}(T)$ \\
\While{$n \notin \mathrm{Leaves}(T)$}{
    \uIf{$\mathrm{Feature}(n) > \mathrm{Threshold}(n)$}{
        $n = \mathrm{Right}(n)$
    }
    \Else{
        $n = \mathrm{Left}(n)$ 
    }
}
$out = \mathrm{Prediction}(n)$
\caption{Decision Tree inference.}\label{alg:dt}
\end{algorithm}

RFs~\cite{breiman2001} are ensembles of DTs (called \textit{weak learners}), each trained differently with a \textit{perturb-and-combine} approach, in order to reduce over-fitting.
Modern RF algorithms~\cite{scikit-learn} perturb weak learners training them: i) on random subsets of the \textit{training samples}, and ii) on random subsets of the input \textit{features}.
At inference time, individual DTs outputs are then combined to obtain the final prediction.
For classification problems, early implementations of RFs let each tree output class labels directly, and the final prediction was computed as the \textit{mode} of such labels. In contrast, in most modern RF libraries~\cite{scikit-learn}, weak learners output the entire array of class probabilities, and the final prediction is computed \textit{averaging} (or equivalently, summing) the probabilities relative to the same class and then taking the ``argmax''.

From a computational standpoint, on a single-core processor, RF inference is therefore implemented with a \textit{sequential loop} over the weak learners. 
This is shown in Algorithm~\ref{alg:rf} for a classification, where $M$ is the number of classes and $\mathrm{DecisionTreeInference}()$ corresponds to Algorithm~\ref{alg:dt}. 
If the $N$ DTs that compose the RF have identical depth $D$, the total memory occupation, as well as the latency and energy consumption due to the main loop will be approximately $N$ times larger than those of a single DT. The time complexity of the argmax, instead, is $O(M)$.

\vspace{-0.2cm}
\begin{algorithm}[ht]
\SetAlgoLined
\small
$out =  \mathbf{0}_{M}$ //array of 0s of size $M$\\
\For{$T \in \mathrm{Forest}$}{
    $out = out + \mathrm{DecisionTreeInference}(T)$
}
$class = \arg\max(out)$
\caption{Random Forest Classification.}\label{alg:rf}
\vspace{-0.1cm}
\end{algorithm}

\section{Related works}

Adaptive inference is increasingly popular for the deployment of energy efficient ML models on IoT end-nodes, as it allows a fine-grained and dynamic control of the trade-off between accuracy and computational complexity~\cite{Daghero2020,Park2015,JahierPagliari2018a,Tann2016,Panda2016,JahierPagliari2020b}.
The basic principle is that not all inputs are equally complex to process (e.g., classify). Easier inputs, which are often the most common, can therefore be treated with a simpler ML model than complex ones or, better yet, involving only a \textit{portion} of the full model. Ideally, this can yield an accuracy comparable to the full model while significantly reducing the average latency and energy consumption.

Several implementations of adaptive inference have been proposed in literature, especially for Deep Neural Networks (DNNs). Among the earliest, Big-Little DNNs~\cite{Park2015} are built combining two networks of different complexity and accuracy. Each input is first fed to the ``little'' (inexpensive and less accurate) model. Then, the confidence of this model's prediction is evaluated, and inference is stopped if such confidence is higher than a threshold. Otherwise, the same input is fed to the ``big'' model. In subsequent works, this idea has been extended to more than 2 networks, and improved using a different confidence threshold per class, or resorting to slimmed-down versions of the ``big'' DNN to implement the ``little'' one(s), e.g., activating only a subset of layers or channels, or applying a lower bit-width quantization~\cite{Tann2016,Panda2016,JahierPagliari2018a,Daghero2020}.

Applications of this paradigm to tree-based learning, and RFs in particular, are less common~\cite{Wang2018,Gao2011,schwing2011}. Moreover, the complexity of the few existing techniques makes them unsuitable for embedded devices. For instance, \cite{Gao2011} proposes two early-stopping criteria, and a mechanism to dynamically select the optimal order of weak learners for each input. The computations involved in this selection, however, have much higher complexity than the evaluation of individual DTs. Hence, as stated in the paper, their approach is only effective if the target task involves complex feature extractions,
and if the latter can be partially skipped in case of early stopping.
The work closest to ours is QWYC~\cite{Wang2018}, which focuses on binary classification problems, and performs early stopping based on two probability thresholds ($\epsilon_{-}$ and $\epsilon_{+}$).
Furthermore, it also statically sorts DTs  so that those with the highest chance of triggering an early stop are invoked first.
During inference, whenever a tree outputs a probability lower than $\epsilon_{-}$ or higher than $\epsilon_{+}$ the execution is stopped, using the negative/positive class as final prediction respectively. While this approach incurs low runtime overheads (two comparisons), the methods used to extract the two thresholds and to sort the weak learners are not easy to extend to multi-class problems.
Importantly, ~\cite{Gao2011},\cite{Wang2018} and~\cite{schwing2011} are only evaluated in terms of complexity reduction, and never deployed on an embedded device to assess their actual energy and latency savings.

%% file: sections/03_method.tex
\subsection{Motivation}
Typical RFs are composed of a large number of trees ($N$), in the order of 10s or even 100s. A large $N$ improves the accuracy of the ensemble, allowing to classify correctly even the hardest input samples. However, for simpler inputs, which are often more frequent, a smaller RF with $N' < N$ trees would suffice to obtain a correct prediction. For those inputs, executing all $N$ weak learners is a waste of time and energy, particularly critical for IoT end-nodes. On the contrary, directly deploying a RF with $N'$ trees could negatively affect the accuracy on difficult samples.

Based on these observations, we devise a simple adaptive \textit{early stopping} policy for RF-based classifiers, which terminates the inference as soon as a high-enough classification confidence is achieved.
The key difficulty in designing such a policy is making it able to accurately distinguish those ``easy'' inputs for which early stopping does not alter the classification result. At the same time, since our main goal is to save energy on resource-limited MCU-based IoT devices, the policy should also introduce minimal runtime overheads with respect to the (lightweight) inference process described in Section~\ref{sec:rf}, to avoid reducing or nullifying the energy savings.

\subsection{Aggregated Score Thresholds for Early Stopping}\label{sec:policy}

The majority of the \textit{lightweight} early stopping policies proposed in previous works~\cite{Park2015,Tann2016,JahierPagliari2018a,Daghero2020,Panda2016,Wang2018} for both deep learning and RF classifiers, is based on comparing a measure of classification confidence with a threshold. Confidence is either measured as the maximum of the class probability scores produced in output by a model - $\max_j(P_j)$ - or as the so-called \textit{Score Margin} (SM), i.e., the difference between the two largest scores:
\begin{equation}
    SM=\max_j(P_j)- \max_{j \ne j_{max}}(P_j) %
\end{equation}
where $j \in [0, M-1]$ is the class index and $j_{max} = \arg\max_j(P_j)$.
Regardless of the specific formula, the confidence measure is typically evaluated on \textit{individual} classifiers. 
For example, in the method of~\cite{Tann2016}, early stopping is triggered whenever $SM_{last} > \alpha$, where $\alpha$ is a tunable threshold, and $SM_{last}$ is the score margin computed on the output probabilities of the \textit{last} classifier executed.

Our work takes inspiration from those policies, but with one significant difference. In fact, we do not decide whether to trigger early stopping based on individual weak learners outputs, but rather, depending on the \textit{aggregated prediction} produced by the part of the ensemble that has been already executed. The rationale for this approach is that,  for easy inputs, the aggregated prediction will be very skewed towards a particular class after a certain amount of trees have been executed, as an effect of all weak learners predicting the correct class with high probability, and of using the sum/mean aggregation described in Section~\ref{sec:rf}. Therefore, it is highly unlikely that the remaining trees will overturn the final decision.

In mathematical terms, we define the partial output of a RF relative to class $j$, after executing $t$ trees as:
\begin{equation}
P^{[0:t-1]}_j=\sum_{i=0}^{t-1}P^{i}_j,\ \forall j \in [0, M-1]
\end{equation}
where $P_j^i$ is the score produced by the $i$-th weak learner for class $j$.
In our experiments, we try two different early stopping policies based either on the largest \textit{aggregated} score $S^{t-1}$ or on the aggregated Score Margin $SM^{t-1}$. In formulas, the first policy stops the inference when: %
\begin{equation}\label{eq:s_t}
S^{t-1} = \max_j(P^{[0:t-1]}_j) > \alpha
\end{equation}
whereas the second one triggers early stopping when the following condition verifies:
\begin{equation}\label{eq:sm_t}
SM^{t-1} = \max_j(P^{[0:t-1]}_j) - \max_{j \ne j_{max}^{t-1}}(P^{[0:{t-1}]}_j) > \alpha
\end{equation}
As shown in Section~\ref{sec:results}, we find $SM^{t-1}$ to be the most reliable confidence metric.
To the best of our knowledge, ours is the first lightweight policy that compares the partially aggregated scores with a threshold to determine early stopping. Nonetheless, our experiments show that this approach is significantly superior to alternatives that base the stopping decision only on the last weak learner executed.

\begin{figure}[ht]
    \centering
    \includegraphics[width=.9\linewidth]{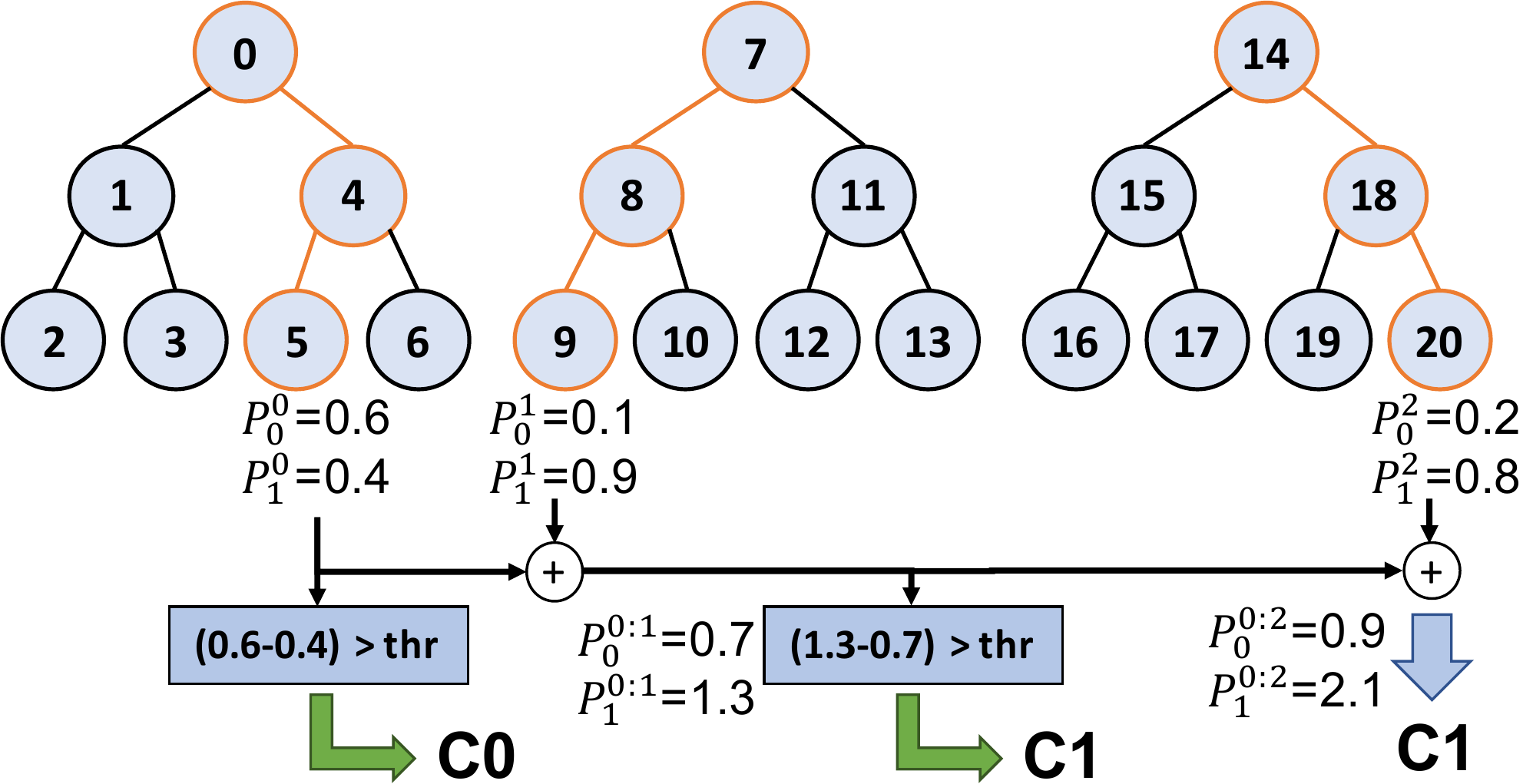}
    \caption{High-level overview of the proposed adaptive inference method for RFs, for a batch of 1. At each step, the SM is computed on the partially aggregated scores.}\label{fig:rf_adaptive}
\end{figure}

A high-level graphical view of our early stopping process is shown in Figure~\ref{fig:rf_adaptive} for $N=3$, $M=2$, $D=3$ and using $SM^{t-1}$ as confidence measure. The orange lines highlight the selected path in each tree. As shown, after the execution of each tree, the partially aggregated scores are computed and then used to decide whether to stop or not.

\subsection{Deployment on MCUs}\label{sec:hw_impl}

The deployment of the proposed adaptive RFs on a MCU requires optimized implementations of: i) the generic RF data structures used for the inference procedure of Algorithm~\ref{alg:rf} and ii) the additional computations required for early stopping.

\subsubsection{Random Forest Implementation}

To the best of our knowledge, there are no open source RF libraries for our target MCU (described in Section~\ref{sec:results}). Therefore, we developed an in-house implementation, inspired by the one available in OpenCV~\cite{opencv_rf}, but optimized for low-memory single-core hardware. Specifically, while OpenCV represents RFs as lists, we employed multiple C arrays to reduce the code size and achieve better memory locality.

\begin{figure}[ht]
    \centering
    \includegraphics[width=.95\linewidth]{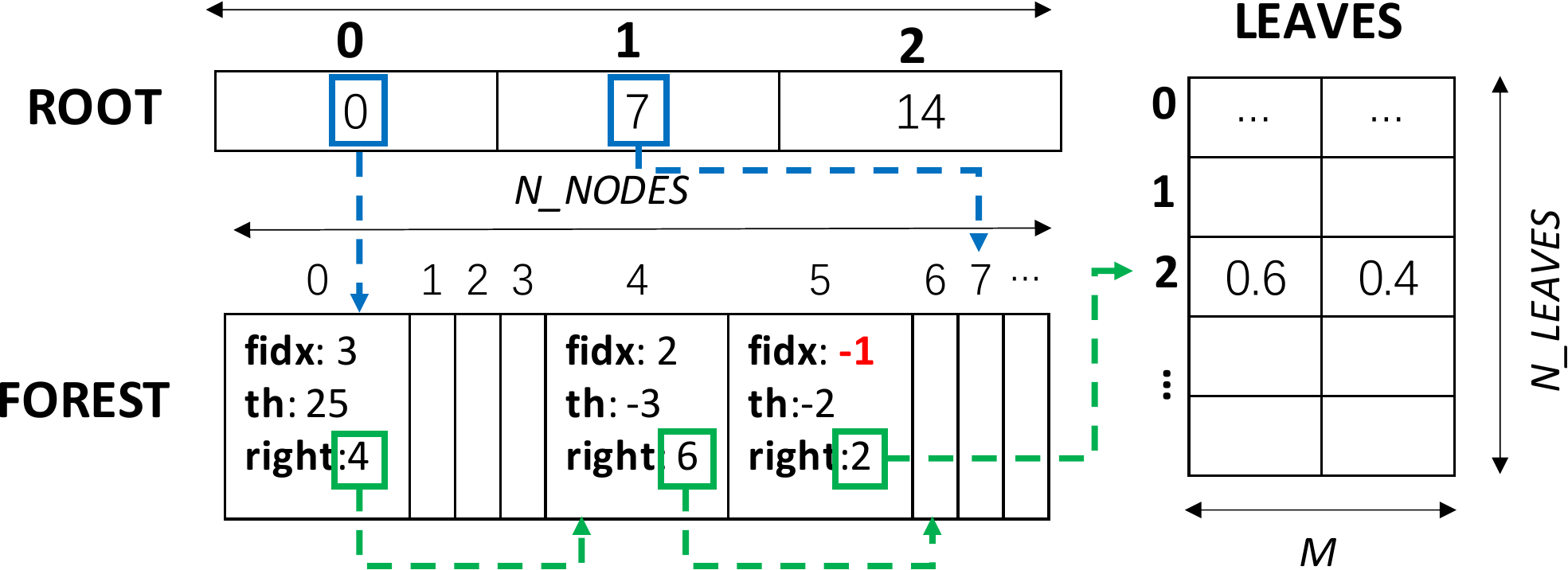}
    \caption{C data structures of our RF implementation.}\label{fig:rf_impl}
    \vspace{-0.3cm}
\end{figure}

Our implementation is based on three main arrays called FOREST, ROOT and LEAVES, as shown in Figure~\ref{fig:rf_impl}.

FOREST is an array of C ``structs'' which stores the details of all nodes. Specifically, each element has three fields:
\begin{itemize}
    \item \textit{fidx}: the index of the input feature that should be compared with \textit{th} in the current node. A special value of -1 is used to identify leaf nodes.
    \item \textit{th}: the comparison threshold used to decide whether to visit the right or left child of the node.
    \item \textit{right}: the index of the right child in FOREST. The left child of node $i$ is implicitly node $i+1$, in order to reduce memory occupation, avoiding an unnecessary \textit{left} field. For the same reason, for leaf nodes, \textit{right} is interpreted as an index in the LEAVES array.
\end{itemize}
LEAVES is a matrix storing the output probabilities for all leaf nodes, while ROOT holds the indexes of the root nodes of the $N$ trees in the RF, and is the starting point for inference. To clarify our implementation, node indexes in Figure~\ref{fig:rf_impl} match those of Figure~\ref{fig:rf_adaptive}. Moreover, the FOREST elements corresponding to the visited nodes of the first tree (nodes 0, 4 and 5) are shown in detail.

Both in the standard RF implementation used as comparison baseline and in our adaptive version, input features, comparison thresholds and output probabilities are \textit{quantized} to 16-bit integers, in order to reduce the memory occupation and inference latency, and to enable deployment on MCUs that do not have a Floating Point Unit. In our experiments, we found that this quantization maintains the same accuracy of the floating point RF. Moreover, index fields are also represented on 16bit, which allows to support RFs with up to $2^{16}$ nodes.

\subsubsection{Early Stopping and Tree Batching}

With respect to the baseline RF implementation, our adaptive version has a negligible memory overhead, as the only additional variable to be stored is the confidence threshold $\alpha$.

Instead, the latency and energy overheads can be significant, despite our simple stopping policy. In fact, assuming to use the SM as confidence measure, deciding whether to stop requires: i) finding the two highest scores in the partially aggregated output, with time complexity $O(M)$ and ii) subtracting them and comparing them with $\alpha$, with complexity $O(1)$. For reference, the evaluation of each DT requires: i) the $O(D)$ visit of the tree, and ii) the $O(M)$ accumulation of its probabilities onto the output array. Therefore, the maximum overheads occur for a classification onto a large number of classes (large $M$) performed with a RF made of shallow trees (small $D$).

We propose a simple and effective way to reduce the overheads due to the evaluation of the early stopping policy based on \textit{tree batching}. Rather than evaluating $SM^{t-1}$ after the execution of each DT, we do so only after a \textit{batch} of $B>1$ weak learners has been evaluated. This has two contrasting effects on latency/energy. On the one hand, it may cause the evaluation of more trees than necessary before early stopping, as the confidence is only evaluated at the end of each batch; on the other hand, it cuts the overheads linked with our policy by a factor $B$. In practice, our experiments show that, depending on the dataset, batching with $B=2$ or $B=4$ may yield superior results than $B=1$.

Figure \ref{alg:rf_adaptive} summarizes our adaptive inference procedure in form of pseudo-code, where $SM()$ is the Score Margin computation and $Batch(b)$ is the set of DTs belonging to the $b$-th batch of size $B$.

\begin{algorithm}[ht]
\SetAlgoLined
\small
\For{$b \in [0, N/B]$}{
    \For{$T \in Batch(b)$}{
        $out= out+ \mathrm{DecisionTreeInference}(T)$
    }
    \If{$SM(out) > \alpha$}{
        break \\
    }
}
$class = \arg\max(out)$
\caption{Adaptive Random Forest Classification.}\label{alg:rf_adaptive}
\vspace{-0.1cm}
\end{algorithm}
\vspace{-0.3cm}

%% file: sections/04_results.tex
\subsection{Setup: Datasets, models and hardware platform}

We test our approach on three datasets of relevant applications.
We use the top-1 accuracy as performance metric if not differently specified. For each dataset, our starting point is a standard (non-adaptive) RF whose parameters ($N$ and $D$) are determined to obtain either i) the smallest model that achieves state-of-the-art accuracy, when possible, or ii) the most accurate model that fits the small memory of the target MCU, otherwise.
For all datasets, we fed the RFs both with raw data or with low complexity time-domain features (feasible to extract on a MCU) proposed in the respective papers. In all cases, we found that raw data yielded the best accuracy, therefore we report the corresponding results.

\begin{figure*}[ht]
\centering
\includegraphics[width=0.88\textwidth]{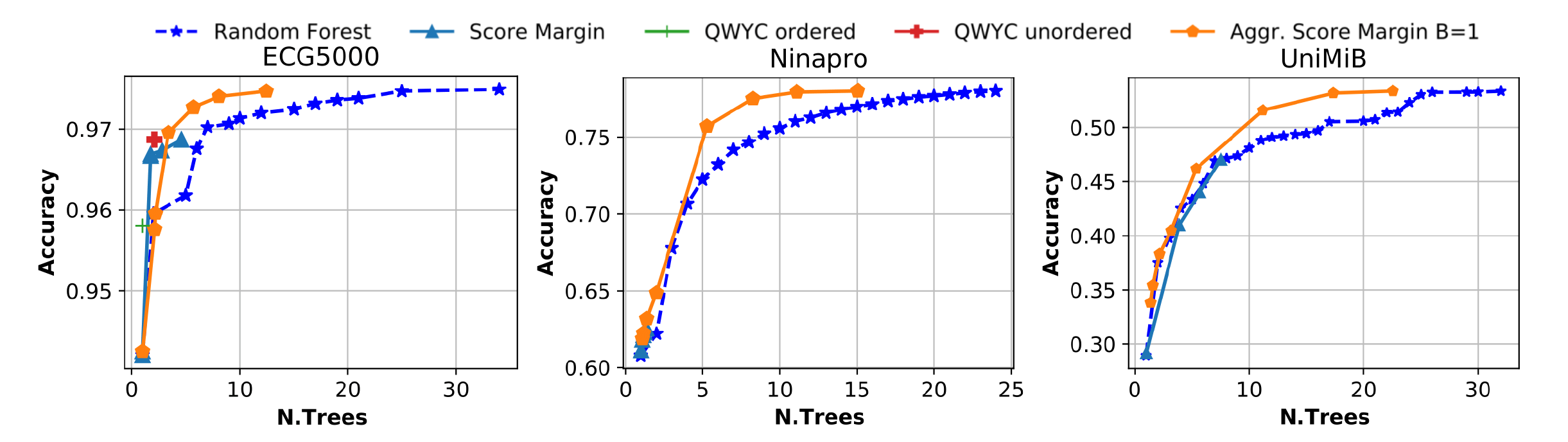}
\caption{Hardware-independent comparison among different adaptive approaches and baseline RFs.}
\vspace{-0.2cm}
\label{fig:accuracy_comparison_trees}
\vspace{-0.4cm}
\end{figure*}

The \textbf{Ninapro DB1}~\cite{atzori2014electromyography} dataset features Electromyography (EMG) signals of multiple hand movements of 27 healthy people. Following the experimental setup proposed by~\cite{atzori2014electromyography}, we perform the classification of 14 hand movements using a 10 channel EMG signal. To classify input samples, we use identical pre-processing (e.g windowing) as in~\cite{atzori2014electromyography} and a random forest with $N=24$ and $D$ = 12 as baseline.
\textbf{UniMiB-SHAR}~\cite{micucci2017unimib} contains tri-axial accelerometer data, used to classify 17 different human activities.  For this dataset, we use a baseline RF with $N=32$ and $D$ = 9. 
Given its class imbalance, we also use the \textit{macro-averaging} accuracy as target metric, as suggested by the authors \cite{micucci2017unimib}.
The \textbf{ECG5000}~\cite{ecg5000} dataset contains an annotated electrocardiogram (ECG) signal of a single patient, divided into 0.8 seconds windows, each containing a single heartbeat. 
We perform a binary anomaly detection on the annotated heartbeats as proposed in \cite{ecgAnomaly}, detecting when a congestive heart failure happens, using a RF with $N=40$ and $D=3$ as baseline.

We deploy all models on the PULPissimo platform, a 32-bit single-core RISC-V MCU,
with 520 KB memory~\cite{quentin}.
We estimate the inference clock cycles using a virtual platform~\cite{gvsoc}, while we derive energy results from the power values of Quentin~\cite{quentin}, a 22nm realization of PULPissimo running at 205 MHz.
The RFs have been trained using the scikit-learn Python package~\cite{scikit-learn}, while MCU code has been written in C.
Besides non-adaptive RFs, we compare our method also with two adaptive solutions: a standard Score Margin evaluated on the last weak learner, such as the one used in~\cite{Park2015,Tann2016,JahierPagliari2018a}, and the QWYC method of~\cite{Wang2018}, which however only supports binary classifications and is therefore only considered for ECG5000. All adaptive methods are applied to a maximally sized RF, with the $N$ and $D$ values reported above.

\subsection{Hardware-independent comparison}

As a first experiment, since none of the state-of-the-art methods for RFs have been deployed on MCUs, we compare them with our proposed approach in a hardware independent way. To this end, we compute the average number of trees executed by each adaptive method when running on the entire validation sets of the three target datasets, for different values of the corresponding early-stopping thresholds ($\alpha$ or $\epsilon_{-}$ and $\epsilon_{+}$). The results are shown in Figure~\ref{fig:accuracy_comparison_trees} as Pareto fronts in the accuracy vs number of trees plane.
For reference, each plot also reports the results of the baseline RF (rightmost blue star), as well as those of smaller (static) RFs built by progressively decreasing the number of trees $N$ (blue dashed Pareto front).

The results show that our Aggregate SM policy can reach an accuracy identical to the baseline RF for all three datasets, while significantly reducing the average number of weak learners executed. Moreover, our approach significantly outperforms a standard SM on the two most difficult datasets (Ninapro and UniMiB). This is due to the fact that the classic SM relies on a single weak learner's prediction, which is often highly confident despite being wrong (except for the easier ECG5000).
The Aggregated SM is also competitive with QWYC on ECG5000, while not being limited to binary problems and allowing a much easier exploration of the accuracy vs complexity space at runtime. In fact, as shown by the single points in the plots, QWYC always yields very similar accuracy and number of trees, regardless of the values assigned to its early-stopping thresholds.

To better quantify the benefits of our approach, Table~\ref{tab:n_trees} reports the average number of trees executed by each method when targeting the same accuracy of the maximally sized (\textit{Full}) RF, or an accuracy drop $<0.5\%$. Besides the Aggregated SM with $B=1$, the table also reports the results obtained with larger batch sizes. The column \textit{Aggr. Max.} shows the results obtained using the maximum aggregated probability as confidence measure (Eq.~\ref{eq:s_t}) instead of the SM, with $B=1$. Lastly, \textit{Red. RF} shows the value of $N$ of a static RF that achieves the same accuracy. The standard SM and QWYC are not reported in the table as they never reach an accuracy with a drop $<0.5\%$ with respect to the full RF.

\input{sections/Tab1}

\subsection{Deployment Results}

\begin{figure*}[ht]
\centering
\includegraphics[width=0.88\textwidth]{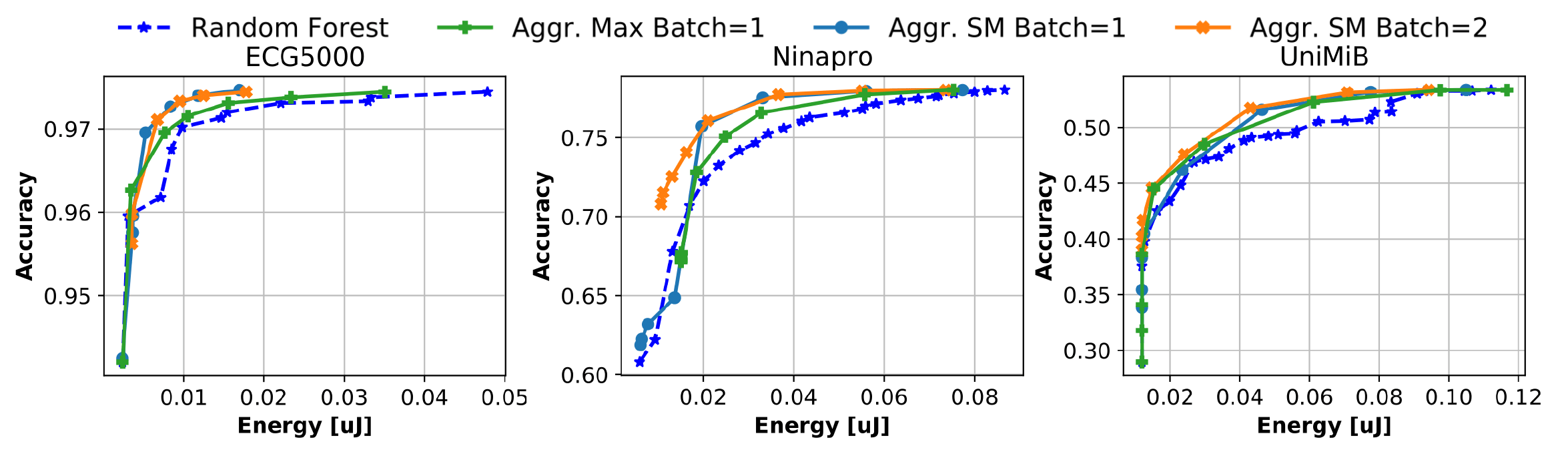}
\caption{Accuracy vs energy consumption of the proposed method for different batches ($B$) and confidence measures.}
\label{fig:accuracy_comparison_energy}
\vspace{-0.4cm}
\end{figure*}

Figure~\ref{fig:accuracy_comparison_energy} and Table~\ref{tab:energy} report the average energy consumption per input obtained deploying the proposed method on PULPissimo. Since ours is the only method deployed on a MCU, the results refer to its different variants.
In particular, we vary again the batch size $B$ (the figure only shows $B=1$ and $B=2$ for clarity)
and consider both the maximum probability score and the Score Margin as confidence measures.

\input{sections/Tab2}

Figure~\ref{fig:accuracy_comparison_energy} shows that the energy vs accuracy trade-off curves have similar shapes to those of Figure~\ref{fig:accuracy_comparison_trees}, thus demonstrating the effectiveness of our method even when taking into account its runtime overheads. Moreover, it shows that using a batch of trees is sometimes preferable (e.g., $B=2$ for the Ninapro dataset). This happens because, while $B=1$ always yields the lowest number of visited DTs, as shown in Table~\ref{tab:n_trees}, the smaller overheads obtained evaluating the policy every $B>1$ trees may outweigh the additional energy costs of visiting a larger number of trees.
This is also the reason why the Aggr. Max policy is considered. In fact, finding a single maximum over the aggregated scores array is slightly more lightweight than computing the SM. However, with this less reliable confidence measure, the visited DTs increase a lot (up to 15 more on average to reach the same accuracy of the Aggr. SM with $B=1$, as shown in Table~\ref{tab:n_trees}). Thus, the overhead reduction is insufficient, and this approach is almost always sub-optimal.

Table~\ref{tab:energy} shows the total energy consumption of different variants of our method, for the same two accuracy conditions of Table~\ref{tab:n_trees}. As anticipated, the lowest energy consumption is often achieved with $B>1$, e.g., on Ninapro ($B=2$) or UniMiB ($B=8$) for a 0\% accuracy drop. Overall, the best variant of our method reduces the energy from a minimum of 15\% (on Ninapro) to a maximum of 65\% (on ECG5000) with no accuracy drop, and from 38\% (UniMiB) to 90\% (ECG5000) when accepting a small accuracy degradation.

%% file: sections/Tab1.tex
\begin{table}[ht]
    \centering
     \caption{Average number of trees for different accuracy drops with respect to a full RF.}\label{tab:n_trees}
    \begin{tabular}{|c|c|c|c|c|c|c|c|}
    \hline 
    \multirow{2}{*}{Data} &  Full & Red. & Aggr. & \multicolumn{4}{|c|}{Aggr. SM}\\\cline{5-8}
    & RF & RF & Max & B=1 & B=2 & B=4 & B=8\\\hline\hline
    \multicolumn{8}{|l|}{Drop: 0\%}\\\hline\hline
     ECG & 40 &40 & 27.17 & \textbf{12.46} & 13.45 & 13.63 & 17.17\\
    Ninapro & 24 & 24 & 16.35 & \textbf{15.05} & 15.66 & 19.31 & 20.71\\
    UniMiB & 32 & 32 & 30.12 & \textbf{22.55} & 22.88 & 23.49 & 24.31 \\\hline\hline
    \multicolumn{8}{|l|}{Drop: 0.5\%}\\\hline\hline
     ECG  &40 & 7 & 5.16 & \textbf{3.40} & 4.52 & 4.66 & 8.09 \\
    Ninapro & 24 &19 & 11.68 & \textbf{8.25} & 9.35 & 10.27 & 10.95 \\
    UniMiB & 32 &25 & 23.81 & \textbf{17.35} & 17.96 & 18.74 & 20.39\\
    \hline
    \end{tabular}
    \vspace{-0.3cm}
\end{table}

%% file: sections/Tab2.tex
\begin{table}[ht]
    \centering
     \caption{Average energy consumption, in $\mu J$, for different accuracy drops with respect to a full RF.}\label{tab:energy}
    \begin{tabular}{|c|c|c|c|c|c|c|c|}
    \hline
    \multirow{2}{*}{Data} &  Full & Red. & Aggr. & \multicolumn{4}{|c|}{Aggr. SM}\\\cline{5-8}
    & RF & RF & Max & B=1 & B=2 & B=4 & B=8\\\hline\hline
    \multicolumn{8}{|l|}{Drop: 0\%}\\\hline\hline
     ECG & 0.048 & 0.048 & 0.035 & \textbf{0.017} & 0.018 & \textbf{0.017} & 0.020 \\
     Ninapro & 0.087 & 0.087 & 0.076 & 0.077 & \textbf{0.074} & 0.083 & 0.083 \\
     UniMiB & 0.112 & 0.112 & 0.117 & 0.105 & 0.094 & 0.090 & \textbf{0.089} \\\hline\hline
    \multicolumn{8}{|l|}{Drop: 0.5\%}\\\hline\hline
    ECG & 0.048 & 0.010 & 0.008 & \textbf{0.005} & 0.007 & 0.006 & 0.010 \\
     Ninapro & 0.087 & 0.071 & 0.050 & \textbf{0.033} & 0.036 & 0.039 & 0.038 \\
     UniMiB & 0.112 & 0.090 & 0.098 & 0.078 & 0.071 & \textbf{0.069} & 0.072 \\
    \hline
    \end{tabular}
    \vspace{-0.3cm}
\end{table}

%% file: sections/05_conclusions.tex
We have presented an adaptive inference method for Random Forests, based on an early stopping policy that reduces the number of weak learners executed for easy inputs. With experiments on three datasets, we have shown that our method performs similarly or better than state-of-the-art adaptive solutions. Moreover, it can be effectively deployed on resource-constrained MCUs, enabling significant energy savings with limited or null accuracy drops. In future work, we will investigate the combination of our method with an optimized ordering of DTs and its extension to multi-core platforms.

%% file: main.bbl
\begin{thebibliography}{10}
\providecommand{\url}[1]{#1}

\bibitem{Samie2019}
F.~Samie \emph{et~al.}, ``{From Cloud Down to Things: An Overview of Machine
  Learning in Internet of Things},'' \emph{IEEE IoT-J},
  vol.~6, no.~3, pp. 4921--4934, 2019.

\bibitem{Zhou2019}
Z.~Zhou \emph{et~al.}, ``{Edge Intelligence: Paving the Last Mile of Artificial
  Intelligence With Edge Computing},'' \emph{Proc. of the IEEE}, vol.
  107, no.~8, pp. 1738--1762, 2019.


\bibitem{Jacob2018}
B.~Jacob \emph{et~al.}, ``{Quantization and Training of Neural Networks for
  Efficient Integer-Arithmetic-Only Inference},'' in \emph{Proc. CVPR}, 2018, pp. 2704--2713.

\bibitem{fan2013}
L.~Fan \emph{et~al.}, ``Human activity recognition model based on decision
  tree,'' in \emph{Proc. CBD}, 2013, p. 64–68.

\bibitem{stsensor}
STMicroelectronics, ``iNemo: always-on 3d accelerometer and 3d gyroscope,'' 2019, \url{www.st.com/resource/en/datasheet/lsm6dsox.pdf}.

\bibitem{Donos2015}
C.~Donos \emph{et~al.}, ``Early seizure detection algorithm based on intracranial eeg and random forest classification,'' \emph{Int. J. Neural Syst.}, vol.~25, no.~05, p. 1550023, 2015.

\bibitem{breiman2001}
L.~Breiman, ``Random forests,'' \emph{Machine learning}, vol.~45, no.~1, pp. 5--32, 2001.

\bibitem{ecgAnomaly}
J. Pereira et al, ``Learning Representations from Healthcare Time Series Data for Unsupervised Anomaly Detection,'' \emph{Proc. BigComp}, 2019, pp. 1-7.

\bibitem{Daghero2020}
F.~Daghero \emph{et~al.}, ``{Energy-Efficient Adaptive Machine Learning on IoT
  End-Nodes With Class-Dependent Confidence},'' \emph{ICECS}, 2020, pp. 1--4.

\bibitem{Park2015}
E.~Park \emph{et~al.}, ``{Big/little deep neural network for ultra low power
  inference},'' in \emph{Proc. CODES+ISSS}, 2015, pp. 124--132. 
  
\bibitem{JahierPagliari2018a}
D.~{Jahier Pagliari} \emph{et~al.}, ``{Dynamic Bit-width Reconfiguration for
  Energy-Efficient Deep Learning Hardware},'' in \emph{Proc. ISLPED}, 2018, pp. 47:1--47:6.

\bibitem{Tann2016}
H.~Tann \emph{et~al.}, ``{Runtime configurable deep neural networks for
  energy-accuracy trade-off},'' in \emph{Proc. CODES}, 2016, pp. 1--10.

\bibitem{Panda2016}
P.~Panda \emph{et~al.}, ``{Conditional Deep Learning for Energy-Efficient and
  Enhanced Pattern Recognition},'' in \emph{Proc. DATE}, 2016, pp. 475--480.

\bibitem{JahierPagliari2020b}
D.~{Jahier Pagliari} \emph{et~al.}, ``{CRIME: Input-Dependent Collaborative
  Inference for Recurrent Neural Networks},'' \emph{IEEE Trans Comput}, 2020.


\bibitem{Gao2011}
T.~Gao \emph{et~al.}, ``{Active Classification based on Value of Classifier},''
  in \emph{Proc. NIPS}, 2011, pp. 1062--1070.

\bibitem{Wang2018}
S.~Wang \emph{et~al.}, ``{Quit When You Can: Efficient Evaluation of Ensembles
  with Ordering Optimization},'' \emph{arXiv Preprint}, abs/1806.11202, 2018.

\bibitem{dt_book}
O.~Z. Maimon \emph{et~al.}, \emph{Data mining with decision trees: theory and
  applications}. World scientific, 2014, vol.~81.

\bibitem{scikit-learn}
F.~Pedregosa \emph{et~al.}, ``Scikit-learn: Machine learning in {P}ython,''
  \emph{J. Mach. Learn. Res.}, vol.~12, pp. 2825--2830, 2011.

\bibitem{opencv_rf}
G.~Bradski, ``{The OpenCV Library},'' \url{ https://opencv.org}, 2000.

  
\bibitem{atzori2014electromyography}
M.~Atzori \emph{et~al.}, ``Electromyography data for non-invasive
  naturally-controlled robotic hand prostheses,'' \emph{Scientific data}, pp. 1--13, 2014.

\bibitem{micucci2017unimib}
D.~Micucci \emph{et~al.}, ``{UniMiB SHAR: A dataset for human activity
  recognition using acceleration data from smartphones},'' \emph{Applied Sciences}, vol.~7, no.~10, p. 1101, 2017.

\bibitem{ecg5000}
Y.~Chen \emph{et~al.}, ``A general framework for never-ending learning from
  time series streams,'' \emph{DMKD}, vol.~29,
  no.~6, pp. 1622--1664, 2015.

\bibitem{quentin}
P.~D. Schiavone \emph{et~al.}, ``Quentin: An ultra-low-power pulpissimo soc in
  22nm fdx,'' in \emph{Proc. S3S}, 2018, pp. 1--3.

\bibitem{gvsoc}
{The PULP Platform}, ``{GVSOC: PULP Virtual Platform},'' 2020. \url{https://github.com/pulp-platform/gvsoc}

\bibitem{schwing2011}
A.~G. Schwing \emph{et~al.}, ``{Adaptive random forest — How many “experts” to ask before making a decision?}``, \emph{CVPR 2011}, 2011, pp. 1377-1384
\end{thebibliography}
